%% file: main.tex
\pdfoutput=1

\documentclass[11pt]{article}

\usepackage[final]{acl}

\usepackage{times}
\usepackage{latexsym}

\usepackage[T1]{fontenc}

\usepackage[utf8]{inputenc}

\usepackage{microtype}

\usepackage{inconsolata}

\usepackage{graphicx}

\usepackage{float}
\usepackage{placeins}
\usepackage{booktabs}
\usepackage{multirow}
\usepackage{tcolorbox}
\usepackage{amsmath}
\usepackage{algorithm}
\usepackage{algorithmic}
\usepackage{enumitem}
\usepackage{balance}
\usepackage{graphicx}
\usepackage{subfigure}
\usepackage{amsmath}
\usepackage{amsthm}
\usepackage{amsfonts}
\usepackage{hyperref}
\usepackage{xspace}
\usepackage{tabularx}
\usepackage{caption}

\title{Benchmarking for Domain-Specific LLMs: A Case Study on \\ Academia and Beyond}

\author{
  \textbf{Rubing Chen}\textsuperscript{\rm 1},
  \textbf{Jiaxin Wu}\textsuperscript{\rm 1},
  \textbf{Jian Wang}\textsuperscript{\rm 1},
  \textbf{Xulu Zhang}\textsuperscript{\rm 1},\\
  \textbf{Wenqi Fan}\textsuperscript{\rm 1},
  \textbf{Chenghua Lin}\textsuperscript{\rm 2},
  \textbf{Xiao-Yong Wei}\textsuperscript{\rm 1, \thanks{Corresponding author}},
  \textbf{Qing Li}\textsuperscript{\rm 1}\\[4pt]
  \textsuperscript{\rm 1}The Hong Kong Polytechnic University \quad
  \textsuperscript{\rm 2}The University of Manchester \\
}

\begin{document}
\maketitle
\begin{abstract}
\input{sections/0-abstract}
\end{abstract}

\input{sections/1-introduction}  
\input{sections/2-related_work}
\input{sections/3-method}
\input{sections/4-benchmark}

\input{sections/5-experiment}
\input{sections/6-conclusion}

\input{sections/limitations}


\bibliography{reference}

\appendix

\input{sections/appendix}

\end{document}

%% file: sections/0-abstract.tex
%
%
%
%
%
%
%
%
%
%

The increasing demand for domain-specific evaluation of large language models (LLMs) has led to the development of numerous benchmarks. These efforts often adhere to the principle of data scaling, relying on large corpora or extensive question-answer (QA) sets to ensure broad coverage.
However, the impact of corpus and QA set design on the \textit{precision} and \textit{recall} of domain-specific LLM performance remains poorly understood.
In this paper, we argue that data scaling is not always the optimal principle for domain-specific benchmark construction. 
Instead, we introduce \textsc{Comp-Comp}, an iterative benchmarking framework grounded in the principle of \textit{comprehensiveness} and \textit{compactness}. 
Comprehensiveness ensures semantic recall by covering the full breadth of the domain, while compactness improves precision by reducing redundancy and noise. 
To demonstrate the effectiveness of our approach, we present a case study conducted at a well-renowned university, resulting in the creation of PolyBench, a large-scale, high-quality academic benchmark.
Although this study focuses on academia, the \textsc{Comp-Comp} framework is domain-agnostic and readily adaptable to a wide range of specialized fields. 
The source code and datasets can be accessed at \textcolor{magenta}{\url{https://github.com/Anya-RB-Chen/COMP-COMP}}.



%% file: sections/1-introduction.tex
\section{Introduction}

\begin{figure}[t!]
    \centering
    \includegraphics[width=1\linewidth]{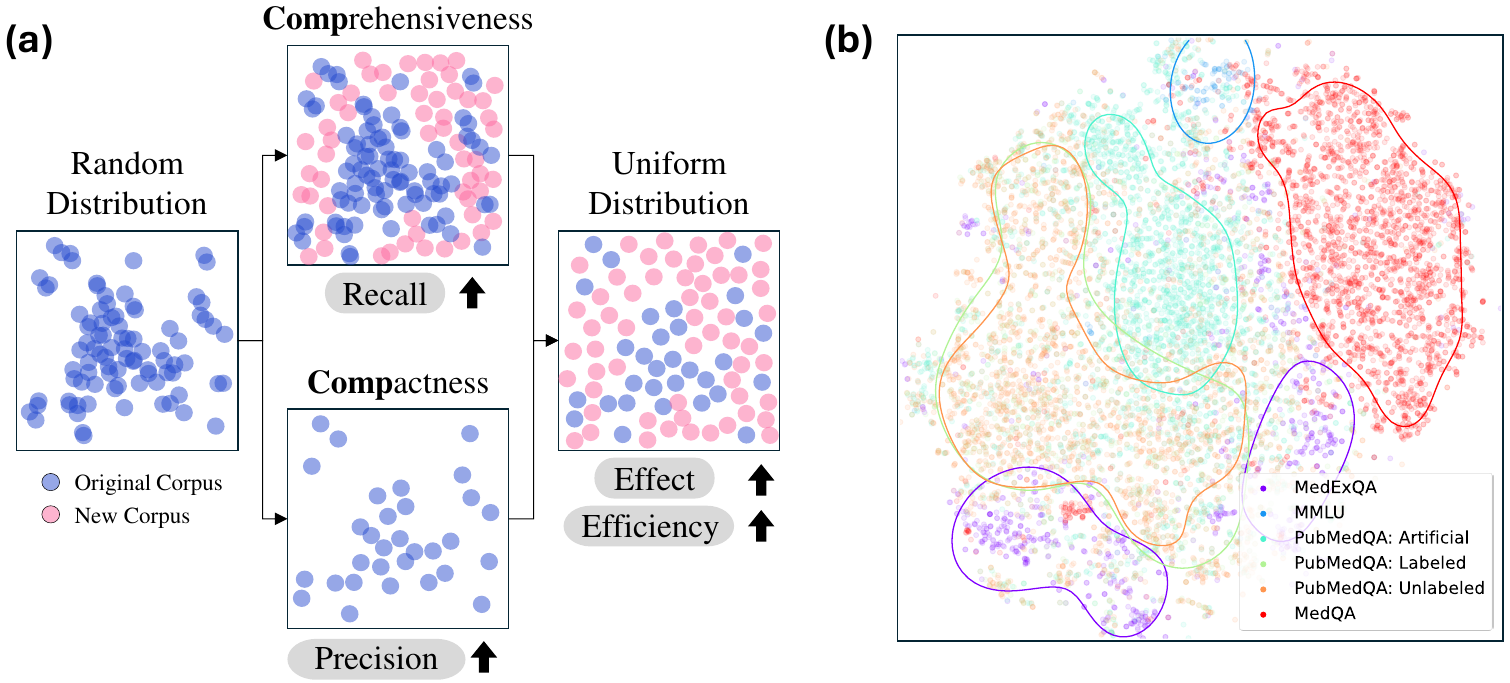}
    \caption{Illustration of \textit{comprehensiveness} and \textit{compactness} in benchmarking. (a) Comprehensiveness ensures broad semantic coverage of the domain, while compactness improves semantic precision by minimizing redundancy, leading to a more uniform data distribution in domain corpora. (b) Semantic distributions of representative benchmarks in the medical domain.}
    \label{fig:compcomp}
\end{figure}

Large language models (LLMs) have demonstrated impressive capabilities across a wide range of tasks \cite{openai2024gpt4technicalreport}. While general-purpose LLMs excel in broad applications, domain-specific LLMs have emerged as critical tools for delivering precise and accurate responses tailored to specific user needs~\cite{fei2023lawbenchbenchmarkinglegalknowledge, Cai2024scieval, chang2024llmevalsurvey}. 
These models are particularly valuable in fields where fixed-answer questions are prevalent, such as law, medicine, and academic, where precision is paramount~\cite{ge2024openagi}. 
However, the development of domain-specific LLMs necessitates high-quality benchmarks that can effectively evaluate their performance. 
Such benchmarks must not only ensure accurate responses within the domain, but also maintain broad coverage of the domain's semantic space to avoid catastrophic forgetting, where models lose proficiency in broader contexts~\cite{luo2023catastrophicforgetting}. 
This dual requirement underscores the need to develop effective approaches for domain-specific benchmarking.

\begin{table*}[t!]
    \centering
    \resizebox{0.82\textwidth}{!}{
    \begin{tabular}{@{}lllll@{}}
    \toprule
    \textbf{Benchmark}  & \textbf{Domain}      & \textbf{Closeness}  & \textbf{QA Type}   & \textbf{\# Questions} \\ \midrule
    SciEval~\cite{Cai2024scieval}      & Science   & Open     & MCQ, QA         & 16,522  \\
    DISC-Law-Eval~\cite{law-disclawllm}   & Law       & Open     & MCQ, MAQ, QA    & 2,863  \\
    Counseling Bench~\cite{med-chatcounselorlargelanguagemodels} & Medical  & Open     & QA              & 229  \\
    CMtMedQA~\cite{med-zhongjing}        & Medical   & Open     & QA              & 70,000  \\
    COVID-QA~\cite{moller-etal-2020-covidqa}       & Healthcare  & Closed    & QA              & 2,019  \\
    WC2014QA~\cite{Zhang2016WC2014QA}        & Sports & Closed   & QA              & 10,211 \\ 
    \midrule
    \textbf{PolyBench (Ours)}   & Academic   &  Closed   & Binary, MAQ, QA, MCQ   & 24,883  \\
    \bottomrule
    \end{tabular}
    }
    \caption{Comparison between domain-specific benchmarks.}
    \label{tab:benchmark_overview}
\end{table*}

Previous studies on constructing domain-specific benchmarks have primarily relied on the principle of data scaling, utilizing extensive corpora or large question sets to ensure broad coverage~\cite{Cai2024scieval, fei2023lawbenchbenchmarkinglegalknowledge}. While this approach is widespread and forms the basis for many benchmark designs, it presents several limitations that can affect the efficacy of model evaluation. First, many benchmarks are derived from skill examinations, practice exercises, or expert-curated datasets, often overlooking a critical assessment of their effectiveness~\cite{med-chatcounselorlargelanguagemodels, ling2023domain}. Although these benchmarks are diverse in question-answering types, ranging from multiple-choice to open-ended formats, they fail to systematically examine the semantic alignment between questions and the underlying corpus~\cite{law-disclawllm, law-chatlaw}. This lack of alignment means that the benchmarks may not fully represent the domain's knowledge structure, leading to evaluations that do not accurately reflect a model's true capabilities within the specific domain.
Second, domain-specific benchmarks often employ LLMs for scoring but neglect to ensure consistency between training and evaluation datasets~\cite{med-alpacare, med-zhongjing}. This oversight can introduce biases where models inadvertently perform better on evaluation tasks due to similarities with the training data, rather than a genuine understanding of the domain. It highlights the need for a more balanced approach to benchmarking, one that integrates both horizontal coverage across a wide range of topics and vertical depth within specific subdomains.

In this paper, we propose \textbf{Comp}rehensiveness-\textbf{Comp}actness (\textbf{\textsc{Comp-Comp}}), a novel framework for constructing domain-specific benchmarks that dynamically balances the semantic distribution of corpora and benchmark questions. As illustrated in Figure~\ref{fig:compcomp}, \textsc{Comp-Comp} iteratively expands the range of corpora and questions, continuously assessing their comprehensiveness (maximizing recall) and compactness (optimizing precision). The framework introduces three crucial mechanisms: 
(1) \textit{Comprehensiveness and Compactness Monitoring}: We encode the corpora and questions into a unified semantic space, using Gaussian Kernel Density Estimation (KDE) to improve these metrics iteratively~\cite{terrell1992variablekde}. 
(2) \textit{Corpora-Question Interaction}: We ensure comprehensive evaluation by generating new questions from underrepresented areas in the corpora. 
(3) \textit{Superiority in Closed Domains}: We prioritize achieving a balance between comprehensiveness and compactness, ensuring the framework excels in constructing closed-domain benchmarks that demand precise corpus evaluation and accurate answer verification.

To validate our framework, we conducted a case study in a well-renowned university and applied \textsc{Comp-Comp} to develop \textbf{PolyBench}, one of the first and most extensive benchmarks for the academic domain (see Table~\ref{tab:benchmark_overview}). PolyBench includes 24.9k questions across various formats and incorporates user-interest-oriented forum discussions, setting a new standard for domain-specific benchmarking. Extensive experiments on the PolyBench demonstrate the practicality and effectiveness of \textsc{Comp-Comp}, providing a scalable approach for future benchmarking in other domains.

In summary, our contributions are as follows: 
(1) We propose \textsc{Comp-Comp}, a principled approach that balances comprehensiveness and compactness in domain-specific benchmark construction. 
(2) We develop PolyBench, a large-scale, high-quality benchmark for the academic domain, which addresses the gap of lacking well-established academic LLM benchmarks. 
(3) Our method is extensible across different domains, offering valuable insights for future benchmarking. By addressing the limitations of existing methods and providing a systematic benchmark design, our work advances the development of domain-specific LLMs.

%% file: sections/2-related_work.tex
\section{Related Work}

\begin{figure*}[t!]
    \centering
    \includegraphics[width=0.9\textwidth]{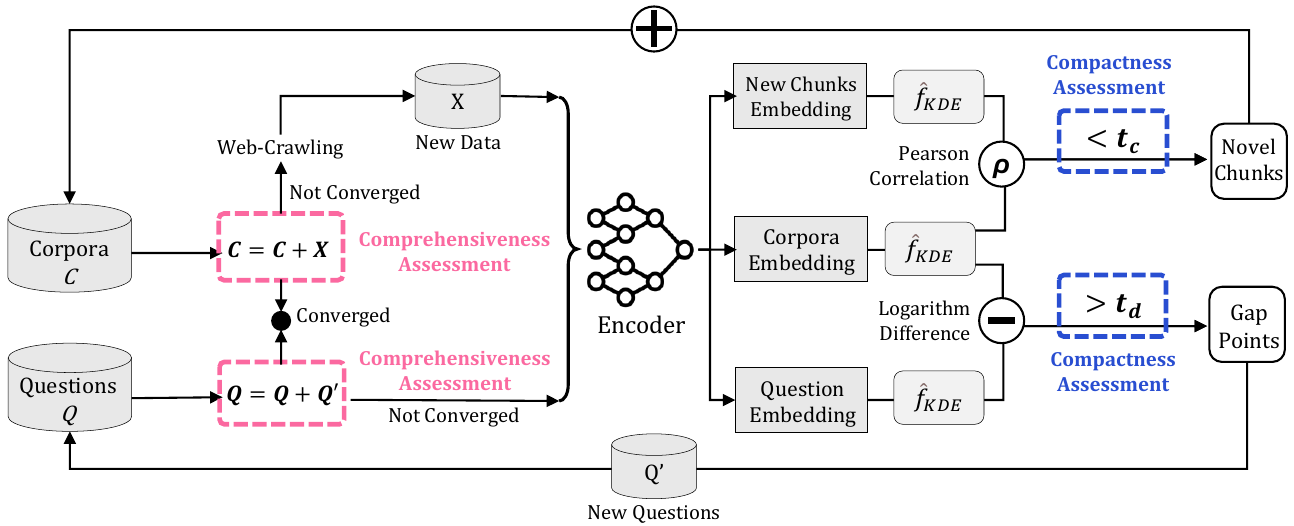}
    \caption{Overview of the proposed Comprehensiveness-Compactness (\textsc{Comp-Comp}) framework for domain-specific benchmark construction. By imposing restrictions represented by $t_c$ and $t_d$, we aim to enhance precision through compactness. Meanwhile, new sets of datasets $X$ or questions $Q'$ are incrementally added to the existing ones, improving comprehensiveness and expanding the semantic coverage.}
    \label{fig:methodology}
\end{figure*}

\paragraph{Domain-Specific Benchmarks}  
Domain-specific evaluation has seen significant advances across critical verticals, each adopting distinct strategies reflective of their unique demands. In the legal domain, systems like DISC-LawLLM \cite{law-disclawllm} employ a dual evaluation paradigm: standardized law examinations (MCQs with tiered difficulty) paired with ChatGPT, arbitrated subjective assessments. LawBench \cite{fei2023lawbenchbenchmarkinglegalknowledge} further refines this by grounding evaluations in Bloom’s Taxonomy \cite{krathwohl2002bloom'sTaxonomy}, systematically testing knowledge memorization, understanding, and application within Chinese legal contexts. The medical domain presents alternative approaches: ChatCounselor \cite{med-chatcounselorlargelanguagemodels} evaluates seven mental consultation dimensions via GPT-4, while CMtMedQA \cite{med-zhongjing} uses dialogue-based capacity assessments scored across three dimensions. Even within single domains, fragmentation persists: Figure~\ref{fig:compcomp}(b) reveals disjointed semantic distributions across medical benchmarks, including MedExQA \cite{kim2024medexqa}, MMLU \cite{hendryckstest2021mmlu}, PubMedQA \cite{jin2019pubmedqa}, and MedQA \cite{jin2021diseasemedqa}, with clear boundaries between corpora despite shared topical grounding. 
Educational benchmarks like C-Eval \cite{huang2024ceval}, though comprehensive, exhibit similar fragmentation, as noted in EduChat’s evaluation \cite{dan2023educhatlargescalelanguagemodelbased}. 
These efforts collectively highlight a critical gap: the absence of unified metrics to assess whether benchmarks sufficiently cover a domain’s semantic space.

\paragraph{Benchmark Construction}  
Contemporary benchmark construction faces three systemic limitations. First, reliance on scaling laws prioritizes quantity over semantic alignment, as seen in legal \cite{law-disclawllm} and medical \cite{med-alpacare} benchmarks built from examination datasets that lack corpus representativeness. Second, LLM-generated evaluations \cite{law-chatlaw,med-zhongjing} risk training-evaluation distribution mismatch, as their metrics often diverge from actual domain usage patterns. Third, concurrence studies like BenchBench \cite{perlitz2024benchbench}, while valuable for assessing general domain robustness, neglect corpus-benchmark integration, a fatal oversight for domain-specific evaluation where training data and test sets are intrinsically linked. These shortcomings collectively manifest in inefficient resource usage: benchmarks either over-collect questions (inflating costs) or under-cover critical domains (compromising validity). It remains underexplored how to maximize the evaluation effect and align the contents between training corpora and benchmarks of the domain-specific LLMs.
To address this, we provide a viable approach and a case study to construct such a benchmark in a systematic scheme.

%% file: sections/3-method.tex
\section{The \textsc{Comp-Comp} Framework}


%

In this section, we present the \textsc{Comp-Comp}, an automatic benchmark construction framework following the comprehensiveness-compactness principle, which guides both the corpora collection and QA generation stages.
Figure~\ref{fig:methodology} shows the overview of \textsc{Comp-Comp} framework.

\begin{figure*}[t!]
    \centering
    \includegraphics[width=1\textwidth]{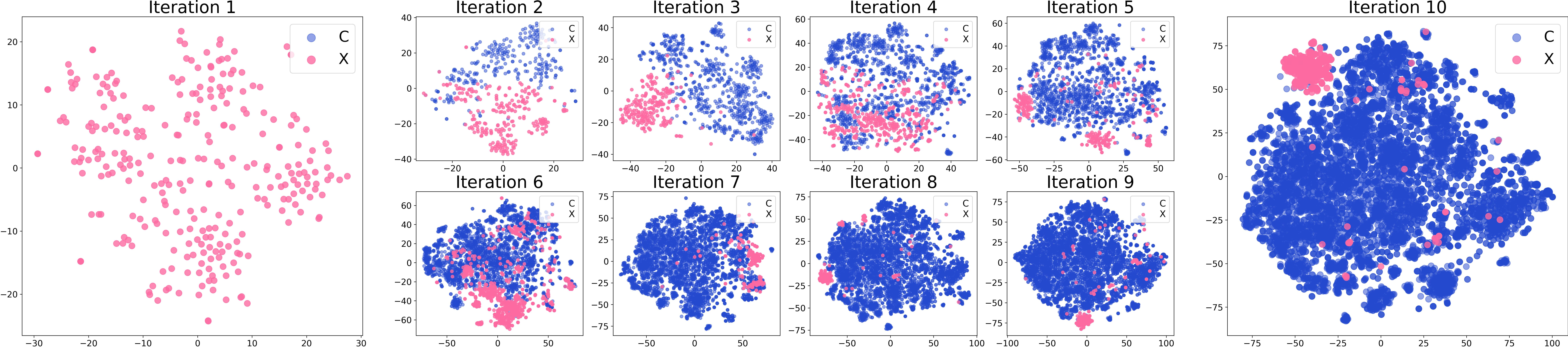}
    \caption{The semantic distribution of 10 sample departments' corpora when collecting Department data from the initial 2k to the final 24k corpora dataset. The pink data points represent the newly added corpora $X$, while the blue points indicate the existing corpora $C$, showcasing the intentional expansion of semantic coverage.}
    \label{fig:deptweb_coverage}
\end{figure*}

\begin{figure*}[th!]
    \centering
    \includegraphics[width=1\textwidth]{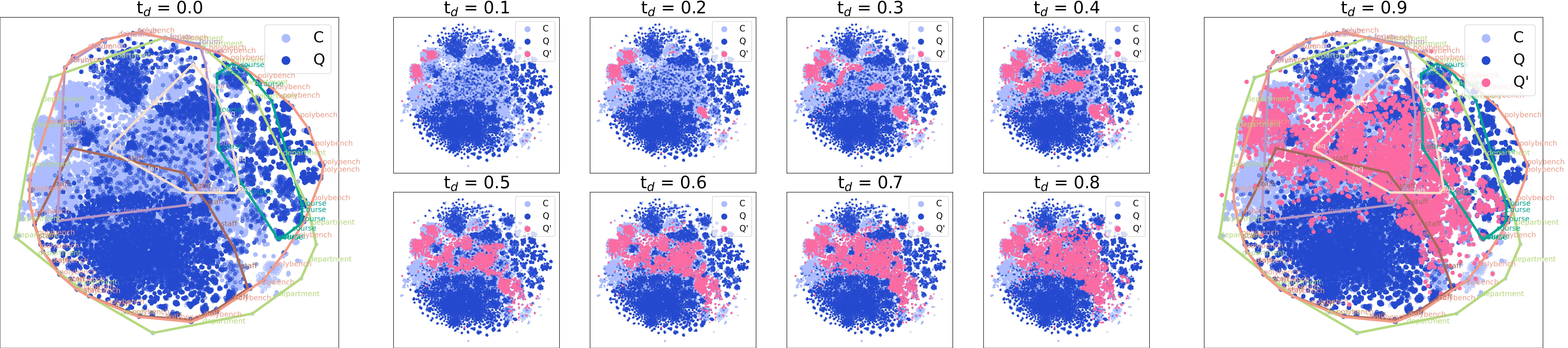}
    \caption{The figure shows how the density threshold $t_d$ influences the semantic coverage of newly added benchmark questions $Q'$. As $t_d$ increases, the gap in coverage of the corpora dataset $C$ by the current benchmark $Q$ will be filled gradually, and the semantic scope of the existing benchmark will be expanded.}
    \label{fig:new_benchmark}
\end{figure*}

\subsection{Comprehensiveness and Compactness Assessment}

Given a set of data point $X = \{x_1,x_2, ..., x_m\}$, its comprehensiveness is measured by computing its coverage of a space $S$ with data points $\{s_1,s_2,...,s_n\}$.
To achieve this, we encode all data points in $X$ and $S$ to the semantic space as follows:
\begin{equation}
\resizebox{\linewidth}{!}{$E_{X} = \{e_{x1},e_{x2},...,e_{xm}\} = \{\text{Encode}(x_i)\,\vert\,\forall(x_i) \in X \}$,}
\end{equation}
\begin{equation}
\resizebox{\linewidth}{!}{$E_{S} = \{e_{s1},e_{s2},...,e_{sn}\} = \{\text{Encode}(s_i)\,\vert\,\forall(x_i) \in S\}$,}
\end{equation}
where $\text{Encode}(\cdot)$ is a text encoder, such as BERT. Then, we use the Gaussian Kernel Density Estimation (KDE), with $j=1,2,...,m$, to estimate their density distributions as follows:
\begin{equation} \label{math:kde1}
    \hat{f}_{\text{X}}(d_j; E_{X}) = \frac{1}{m h} \sum_{i=1}^{m} \exp\left( -\frac{\| d_j - e_{xi} \|^2}{2h^2} \right),
\end{equation}
\begin{equation} \label{math:kde2}
    \hat{f}_{\text{S}}(d_j; E_{S}) = \frac{1}{n h} \sum_{i=1}^{n} \exp\left( -\frac{\| d_j - e_{si} \|^2}{2h^2} \right).
\end{equation}
Based on the distributions, we can estimate the points in the non-overlap part (denoted as gap points) by capturing how much more (or less) dense the $S$ is relative to the $X$ at each point.
This is by calculating their logarithm of density ratio as:
\begin{equation}
    \Delta \log f(d_j) = \log \left( \frac{\hat{f}_{\text{S}}(d_j)}{\hat{f}_{\text{X}}(d_j)} \right),
\end{equation}
and we have non-overlap area of the set $S$ relative to the space $X$ as 
$A_{gap}(X,S) = \{ d_j|\Delta \log f(d_j) > 0, d_{j}\in E_S\}$. 

On the other hand, the compactness of the corpora and QA is controlled by whether the inclusion of new data points $X$ gives much semantic residency to the existing set $Y$. Thus, we measure the Pearson correlation coefficient~\cite{cohen2009pearsonpcc} between the semantic distributions of $X$ and $Y$ as:
\begin{equation}
    r(X,Y)= \rho(\hat{f}_{X}(d_j),\hat{f}_{Y}(d_j)),
\end{equation}
where $\rho(\cdot)$ is the correlation metric. If their coefficient $r(X,Y)$ is lower than a threshold $t_c$, it is considered the inclusion of $X$ to $Y$ will not affect the compactness of $Y$.

\subsection{Assessment Guided Iterative Corpora Expansion}
After the data crawler of domain-related material as $S$, we need to process the data chunks in $S$ to construct a corpora $C$.
The process ensures that the corpora cover as much semantic space of $S$ as possible to obtain good comprehensiveness, while the redundancy in the corpora should be low to achieve compactness.
We formulate corpora processing as a semantic expansion and completion process.   
%
A subset of available data $X\in S$ is added to the current corpora only if it can fill the semantic gap and expand the boundaries of semantic knowledge current $C$.
The whole process is achieved by the following iterative function:
\begin{equation} \label{math:corpora}
    C = C + X, \text{if }  r(X,C) < t_c, \forall X \in S.
\end{equation}
Using this function, we iteratively add a new subset $X \in S$ to $C$ if the inclusion does not introduce many redundancy to the existing corpora set.
Meanwhile, we also observe the non-overlap area $A_{gap}(C,S)$ to monitor the comprehensiveness.
The iteration stops if there is no more $X \in S$ and no gap points in $A_{gap}(C,S)$. 
%
%
%
%

\begin{figure*}[t!]
    \centering
    \includegraphics[width=0.9\linewidth]{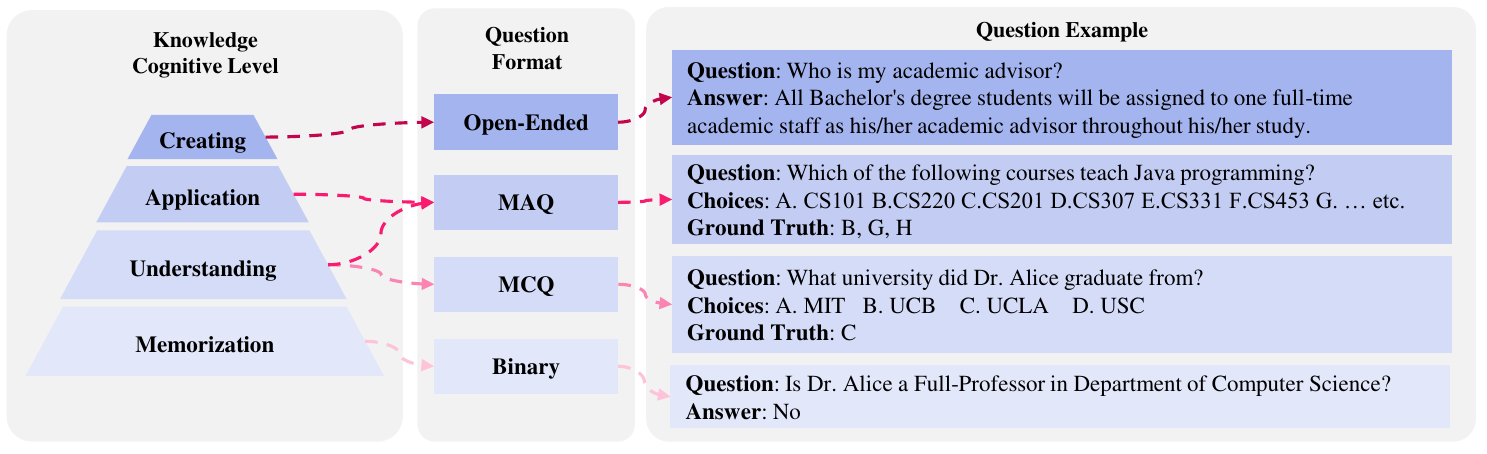}
    \caption{Example questions for corresponding question formats and cognitive levels.}
    \label{fig:polybench_examples}
\end{figure*}


\subsection{Mining User Interests for QA Generation}

We follow the comprehensiveness-compactness principle in the QA generation process. 
After having the processed corpora $C$, our aim here is to generate a QA set $Q$ with diverse content to cover the whole semantic space of the corpora to achieve comprehensiveness, and the generated questions should be representative to ensure compactness.
%
%

To this end, we propose a QA generation workflow as a loop process, as shown in Figure \ref{fig:methodology}.
In the first round, the question set $Q$ is initialized by generating QAs based on the whole corpora using a QA generation function $f_{QAGen}(C)$. 
The QA generation function $f_{QAGen}$ accepts corpora $C$ as input and uses scheme QA generation method for structured data chunks, such as SQARQL for data with Ontology structured and uses LLM to generate QA for unstructured data chunks. 
The questions cover different knowledge cognitive levels, such as creation, application, understanding, memorization, and different formats, such as open-ended, MQA, MCQ, Binary questions.
%
%
After the first round, we obtain the non-overlapped semantic area $A_{gap}'(C,Q)$ between the corpora $C$ and the current QA set $Q$. 
Using a density threshold $t_d$ from zero to one, we can decide to what extent the gaps between the corpora $C$ and the QA set $Q$ are filled. $t_d$ is set based on a specified percentile of the density differences, and this percentile determines the cutoff point for what is considered a ``significant'' density difference.
\begin{equation} \label{math:density}
    A_{gap}'(C,Q) = \{ d_j \mid \Delta \log f(d_j) > t_d ,  d_j \in E_{Q} \},
\end{equation}
\begin{equation}
   Q = Q + f_{\text{QAGen}}\left(  A_{gap}'(C,Q) \right).
\end{equation}
The $A_{gap}'(C,Q)$ represents the semantic area that we aim to cover in the next round of QA generation. 
New QA set $Q'=f_{\text{QAGen}}\left(A_{gap}'(C,Q) \right)$, can be generated using the QA generation function $f_{QAGen}$ on the targeting corpora  $A_{gap}'(C,Q)$.

Moreover, in the QA generation, we propose that we should not only generate QAs based on the corpora $C$ but also include user-interested questions. For example, the frequently asked questions and the questions discussed in public forums (e.g., Quora
and Zhihu).
To this end, we add user-interested questions as a special gap points to $A_{gap}'(C,Q)$.
%




%% file: sections/4-benchmark.tex
\section{PolyBench: An Academic Benchmark}
\label{sec:benchmark_data}
In this section, we employ the proposed framework to construct \textbf{PolyBench}, a comprehensive and compact benchmark in the academic domain.

\subsection{Assessment Guided Corpora Collection}


The \textsc{Comp-Comp} framework governs corpus construction through iterative semantic distribution analysis, as formalized by $C_{i+1} = C_i \cup f_{Corpus}(X, A_{gap}(C_i, S))$, where $A_{gap}$ identifies undercovered semantic regions. Figure~\ref{fig:deptweb_coverage} demonstrates this process through three critical phases. Our dual-phase validation ensures both dimensions of \textsc{Comp-Comp}:
(1) \textbf{Comprehensiveness}: Semantic coverage expands 3.75$\times$ (20$\times$20 to 75$\times$75) through gap analysis, capturing 98\% of academic domain concepts in terms of staff and courses.
(2) \textbf{Compactness}: The final corpora eliminate 68\% of redundant entries compared to unfiltered web crawls, verified through pairwise similarity analysis (cosine distance $<$ 0.2 in 92\% of cases).

\begin{table}[t!]
    \centering
    \resizebox{1.0\linewidth}{!}{
    \begin{tabular}{@{}llllr@{}} 
        \toprule
        \textbf{Cog.} & \textbf{Type} & \textbf{Source} & \textbf{Metric} & \textbf{\# QA} \\
        \midrule
        \multirow{2}{*}{KM} & Binary & Staff & Accurary & 5,000 \\
                            & Binary & Course & Accurary & 2,000 \\
        \midrule
        \multirow{3}{*}{KU} & MCQ & Staff & Accurary  & 5,000 \\
                            & MAQ & Staff & Recall, Precision, F1  & 1,000 \\
                            & MCQ & Course & Accurary  & 2,000 \\
        \midrule
        \multirow{2}{*}{KA} & MAQ & Staff & Recall, Precision, F1  & 2,898 \\
                            & MAQ & Course & Recall, Precision, F1  & 1,598 \\
        \midrule
        \multirow{3}{*}{KC} & Open-ended & Department & BLEU-2, BLEU-4  & 2,726 \\
                            & Open-ended & FAQ & BLEU-2, BLEU-4  & 902 \\
                            & Open-ended & Forum & BLEU-2, BLEU-4  & 1,759 \\
        \bottomrule
    \end{tabular}
    }
    \caption{Statistics of the PolyBench.}
    \label{tab:benchmark_stat}
\end{table}

\subsection{Assessment and Interest Guided QA Construction}


Our QA generation methodology operates through an iterative threshold guided process, formalized as $Q_{i+1} = Q_i \cup f_{QAGen}(A_{gap}'(C,Q_i), t_d)$, where $A_{gap}'$ denotes the non-overlapped semantic area between corpus $C$ and current QA set $Q_i$. For structured corpora (staff/course schemas), we implement automated template filling for binary, MCQ, and MAQ formats using regular expression patterns and SPARQL \cite{cyganiak2005sparqlrelational} based ontology reasoning. Unstructured text (department documents) undergoes LLM-driven reading comprehension QA extraction via LLaMA3-8B, while FAQ/Forum data leverages inherent Q\&A structures. More implementation details are presented in Appendix~\ref{sec:app_qagen}.

The density threshold $t_d$ governs iterative benchmark expansion through gap analysis, visualized in Figure~\ref{fig:new_benchmark}. Higher $t_d$ values enforce tighter semantic coverage, with light/dark blue dots representing $C$ and $Q_i$, and claret-red dots showing new $Q'$ from $A_{gap}'$. This achieves comprehensive coverage with minimal questions—final benchmarks require only 1.7\% of potential QA pairs (24,883 total) while maintaining domain representativeness (Table~\ref{tab:benchmark_stat}).

%% file: sections/5-experiment.tex
\section{Experiments}

\begin{table*}[t!]
\small
\centering
\renewcommand{\arraystretch}{0.85}
\resizebox{1.0\textwidth}{!}{
\begin{tabular}{@{}cllcccccccccc@{}} 
\toprule
\multirow{2}{*}{\textbf{Paradigm}} & \multirow{2}{*}{\textbf{LLM}} & \multirow{2}{*}{\textbf{Method}}  & \textbf{Binary}    &    \textbf{MCQ}     & \multicolumn{3}{c}{\textbf{MAQ} (on understanding)}      & \multicolumn{3}{c}{\textbf{MAQ} (on  application)}    & \multicolumn{2}{c}{\textbf{Open-ended}} \\
\cmidrule(lr){4-4} \cmidrule(lr){5-5} \cmidrule(lr){6-8} \cmidrule(lr){9-11} \cmidrule(lr){12-13}
&         &         & Acc.  & Acc. & Recall & Precision & F1 & Recall & Precision & F1     & BLEU-2 & BLEU-4 \\
\midrule
\multirow{9}{*}{ICL} &
\multirow{3}{*}{LLaMA3-8B} & Vanilla & 0.525 & 0.354 & 0.473 & 0.546 & 0.474 & 0.665 & 0.512 & 0.541 & 0.191 & 0.126 \\
             && Fewshot & 0.534 & 0.392 & 0.410 & 0.534 & 0.440 & 0.492 & 0.532 & 0.475 & 0.227 & 0.153 \\
             && RAG     & \textbf{0.893} & 0.758 & 0.903 & 0.806 & 0.820 & 0.603 & 0.558 & 0.533 & 0.345 & 0.298 \\
\cmidrule(lr){2-13}
& \multirow{3}{*}{GLM4-9B} & Vanilla & 0.525 & 0.364 & 0.739 & 0.530 & 0.580 & \textbf{0.865} & 0.500 & 0.599 & 0.190 & 0.123 \\
             && Fewshot & 0.527 & 0.447 & 0.624 & 0.578 & 0.574 & 0.658 & 0.572 & 0.581 & 0.196 & 0.128 \\
             && RAG     & 0.834 & 0.776 & \textbf{0.964} & 0.757 & 0.826 & 0.696 & 0.530 & 0.553 & 0.309 & 0.256 \\
\cmidrule(lr){2-13}
& \multirow{3}{*}{GPT-4} & Vanilla & 0.694 & 0.568 & 0.641 & 0.650 & 0.613 & 0.724 & 0.661 & 0.652 & 0.291 & 0.201 \\
             && Fewshot & 0.704 & 0.633 & 0.713 & 0.642 & 0.647 & 0.745 & 0.722 & \textbf{0.702} & 0.314 & 0.220 \\
             && RAG     & 0.792 & \textbf{0.783} & 0.932 & \textbf{0.975} & \textbf{0.941} & 0.637 & \textbf{0.807} & 0.670 & 0.364 & 0.315 \\
\midrule
\multirow{9}{*}{SFT} &
\multirow{3}{*}{LLaMA3-8B} & Vanilla & 0.552    & 0.404   & 0.559 & 0.545    & 0.513 & 0.675 & 0.533    & 0.559 & 0.280 & 0.193 \\
             && Fewshot & 0.536    & 0.393   & 0.492 & 0.525    & 0.473 & 0.571 & 0.532    & 0.514 & 0.285 & 0.196 \\
             && RAG     & 0.791    & 0.803   & 0.830 & 0.874    & 0.820 & 0.587 & 0.576    & 0.528 & 0.427 & 0.376 \\
\cmidrule(lr){2-13}
& \multirow{3}{*}{GLM4-9B} & Vanilla & 0.532    & 0.440   & 0.752 & 0.567    & 0.610 & 0.770 & 0.536    & 0.600 & 0.279 & 0.196 \\
             && Fewshot & 0.500    & 0.468   & 0.799 & 0.531    & 0.600 & 0.845 & 0.517    & 0.606 & 0.285 & 0.201 \\
             && RAG     & \textbf{0.840}    & \textbf{0.900}   & \textbf{0.942} & \textbf{0.959}    & \textbf{0.940} & 0.694 & \textbf{0.684}    & \textbf{0.642} & \textbf{0.442} & \textbf{0.397} \\
\cmidrule(lr){2-13}
& \multirow{3}{*}{Qwen2-7B} & Vanilla & 0.608    & 0.402   & 0.825 & 0.500    & 0.590 & 0.827 & 0.521    & 0.606 & 0.263 & 0.178 \\
             && Fewshot & 0.587    & 0.392   & 0.811 & 0.453    & 0.548 & \textbf{0.858} & 0.481    & 0.576 & 0.259 & 0.177 \\
             && RAG     & 0.674    & 0.830   & 0.852 & 0.848    & 0.821 & 0.623 & 0.559    & 0.532 & 0.398 & 0.351 \\
\bottomrule
\end{tabular}
}
\caption{The performance of In-Context Learning (ICL) and Supervised Fine-tuning (SFT) approaches (i.e., few-shot learning and RAG) on the PolyBench with various types of questions.}
\label{tab:results}
\end{table*}


\begin{table*}[th!]
\centering
\small
\centering
\resizebox{1.0\textwidth}{!}{
\begin{tabular}{@{}lccccccccccccccc@{}}
\toprule
\multirow{2}{*}{\textbf{Model}} & \multicolumn{6}{c}{\textbf{$t_d$ thresholds for questions $Q$}} && \multicolumn{8}{c}{\textbf{$t_c$ thresholds for corpus $C$}} \\
\cmidrule(lr){2-7} \cmidrule(lr){9-16}
 & w/o $t_d$ & $0.1$ & $0.3$ & $0.5$ & $0.7$ & $0.9$ && w/o $t_c$ & $0$ & $0.05$ & $0.1$ & $0.15$ & $0.2$ & $0.25$ & $0.3$ \\
\midrule
  & 24591 & \textbf{424} & 1189 & 2172 & 3247 & 4070  & &  25376 & \textbf{11776} & 12252 & 16770 & 18714 & 19498 & 20821 & 21821 \\ \midrule
Vicuna-7B & 0.084 & \textbf{0.060} & 0.082 & 0.087 & 0.086 & 0.086      && 0.067 & 0.064 & 0.067 & 0.069 & 0.068 & 0.129 & 0.071 & \textbf{0.131} \\
Vicuna-13B & 0.092 & \textbf{0.063} & 0.077 & 0.087 & 0.087 & 0.087     && 0.066 & 0.067 & 0.072 & 0.119 & 0.134 & 0.143 & 0.147 & \textbf{0.147} \\
Llama3-8B & 0.374 & \textbf{0.363} & 0.376 & 0.389 & 0.389 & 0.390      && 0.296 & 0.296 & 0.312 & 0.302 & 0.317 & 0.318 & \textbf{0.320} & 0.317 \\
Qwen2-7B & 0.295 & 0.299 & \textbf{0.298} & 0.300 & 0.302 & 0.301       && 0.183 & 0.182 & 0.174 & 0.187 & 0.189 & \textbf{0.190} & 0.190 & 0.160 \\
GLM4-9B & 0.360 & \textbf{0.339} & 0.350 & 0.351 & 0.354 & 0.353        && 0.247 & 0.245 & 0.247 & 0.250 & \textbf{0.257} & 0.166 & 0.257 & 0.172 \\
\bottomrule
\end{tabular}
}
\caption{The performance of frozen models on a subset of the Open-Ended questions with RAG baseline across different thresholds $t_d$ and $t_c$, parameterize the distribution of questions $Q$ and corpus $C$ respectively.}
\label{tab:ablation}
\end{table*}

\subsection{Experimental Settings}

\paragraph{Baselines}
We include both frozen and supervised fine-tuning (SFT) LLMs with in-context learning (ICL) prompts as baselines.
For the ICL, we apply few-shot learning \cite{parnami2022fewshotlearning} and Retrieval Augmented Generation (RAG) \cite{lewis2020naiveRAG, gao2023RAGsurvey} to include domain knowledge in the prompt for LLMs.
%
For the SFT, we use LoRA \cite{hu2021lora} to fine-tune several LLM backbones. 
%
%
%
For LLM backones, we apply both close-sourced and open-sourced LLMs including GPT3.5, GPT4 \cite{openai2024gpt4technicalreport}, Vicuna-7B, Vicuna-13B \cite{vicuna2023}, LLaMA3-8B \cite{touvron2023llama}, GLM4-9B \cite{glm2024chatglm}, and Qwen2-7B \cite{qwen}.

\paragraph{Implementation Details}
For the PolyBench construction, we set the key hyperparameters based on preliminary experiments aimed at balancing comprehensiveness and compactness. The corpus compactness threshold was set to $t_c=0.05$ in Eq. (\ref{math:corpora}) for the corpora collection and $h=5.0$ in Eq.(\ref{math:kde1}) and Eq. (\ref{math:kde2}), and $t_d=0.6$ in Eq (\ref{math:density}) for the QA generation. Results of more parameters and their sensitivity will be presented in Section~\ref{sec:ablation}.
We use Gte-Large \cite{li2023gte-large-embedding} encoder to encode data chunks, selected for its robust performance in capturing nuanced semantic similarities.



\subsection{In-Context Learning on PolyBench}



Table \ref{tab:results} compares the performances of different in-context-learning techniques, i.e., few-shot learning and RAG, on answering several different QA sets of PolyBench. Experiment results on more LLMs are listed in Appendix \ref{sec:app_experiment_results}'s Table \ref{tab:result_ic}. Based on the results, we have several findings:


\paragraph{RAG Enhances Performance Across Most Question Types.} Our analysis demonstrates that RAG substantially improves precision and recall over vanilla and few-shot learning in domain-specific QA. This holds consistently across major backbone models and question types, with performance gains ranging from 2\%--86\% over vanilla baselines and 2\%--88\% over few-shot baselines. However, RAG exhibits a precision-recall tradeoff for complex reasoning MAQs, achieving higher precision but lower recall across all main models. This suggests benchmarks should explicitly account for task priorities (precision-oriented vs. recall-oriented scenarios) when evaluating QA systems.

\paragraph{Few-Shot Learning Suffers from Answer Diversity.} While few-shot learning marginally improves binary and MCQ responses by clarifying answer formats, 33\% of its implementations underperform vanilla prompts. All main models except GPT-4 show metric degradation in $\geq$2 categories, with MAQs being most vulnerable: LLMs often misinterpret few-shot examples as definitive answer templates. Appendix \ref{sec:app_experiment_results} data reveals extreme cases (e.g., Vicuna-13B's 90\% metric decline), highlighting the risk of overfitting to limited examples. This implies that few-shot strategies require careful validation against answer diversity.

\subsection{Supervised Fine-tuning on PolyBench}

The results of the SFT models are listed in Table \ref{tab:result_ft}. 
Compared to the vanilla results of the frozen model in Table \ref{tab:result_ic}, we can find that fine-tuning techniques can consistently improve recall and precision across different types and difficulties of domain-specific questions for the latest LLMs, such as LLaMA3, GLM4, and Qwen2. 
The average improvement from frozen LLMs to SFT LLMs with vanilla prompts is around 44.3\% at each metric.
Moreover, the SFT prominently improves open-ended questions across different LLM backbones.
%
%
The average improvement of all models on open-ended questions is 131.84\% on BLEU-2 and 163.71\% on BLEU-4. 
%
The largest improvement is with the Vicuna models. Vicuna-7B experiences a 194.2\% and 236.11\% increase in average BLEU-2 and BLEU-4, while Vicuna-13B sees a 347.5\% and 442.7\% increase, respectively.
For the other models, the average increase is at least 32.19\% increase on Open-Ended questions after fine-tuning.
Such boosting is noteworthy. We can gain insights that a benchmark with diverse question formats matters if we want to evaluate the model thoroughly.

We also conduct the experiments by incorporating the in-context-learning technique (e.g., few-shot learning and RAG) into SFT LLMs. 


%

Among all the results of SFT in Table \ref{tab:result_ft}, 30.67\% experienced worse results than the frozen LLMs in Table \ref{tab:result_ic} under the same settings.
%
The majority of them are from MAQ questions, which require more reasoning ability than other question formats.
This phenomenon indicates that even though the SF models gain domain knowledge, they suffer from catastrophic forgetting, which downgrades their general capabilities in handling long context information and reading comprehension.
This could still be an ongoing issue in the current research field regarding how to balance domain-specific knowledge and general abilities.

\begin{figure}[t!]
    \centering
    \includegraphics[width=1\linewidth]{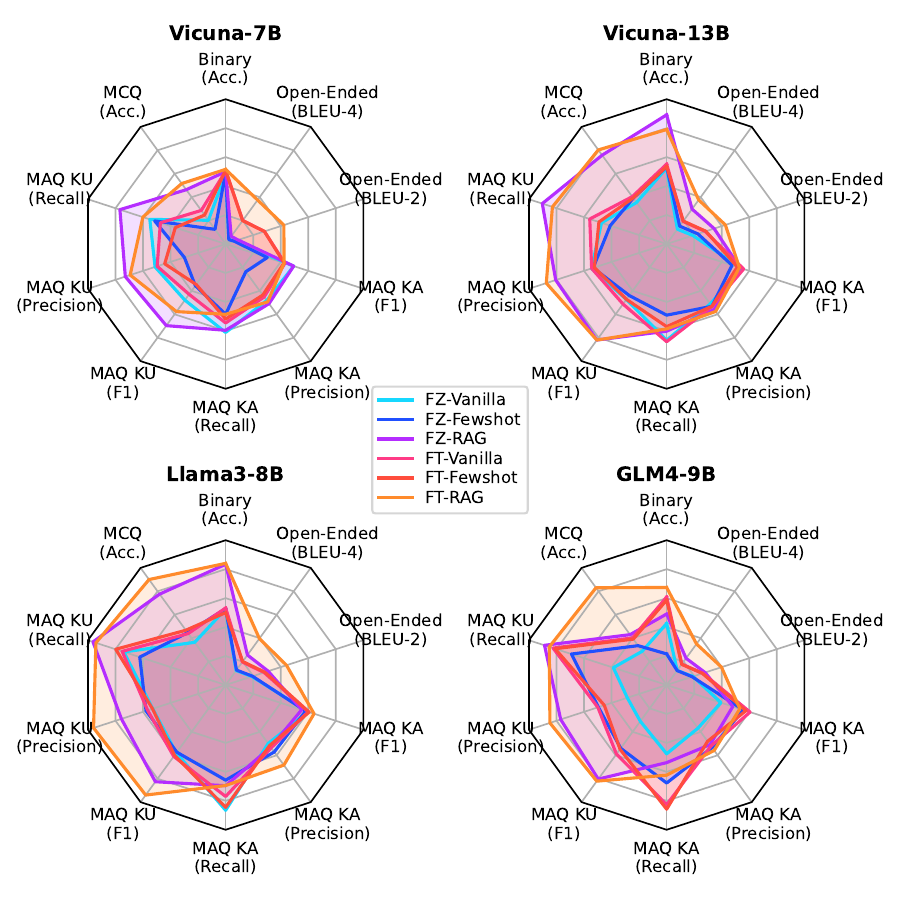}
    \caption{The radar chart demonstrates the collaboration of Frozen models (FZ) and Fine-Tuning (FT) techniques. Each sub-chart has the results of using frozen and fine-tuning models with three types of prompts: Vanilla, Few-shot Learning, and RAG.}
    \label{fig:ic_ft}
\end{figure}

\subsection{Comparison between ICL and SFT}
We also visualize the comparison of ICL and SFT techniques in Figure \ref{fig:ic_ft}. In each sub-figure, two groups of results are compared, i.e., frozen (namely FZ) and fine-tuning LLM (namely FT). Each group contains three results obtained by vanilla, few-shot, and RAG prompting techniques. 

The results show that the RAG technique with a fine-tuning model yields the best precision and recall in answering domain-specific questions.
For example, using RAG with fine-tuned LLaMa3-8B, around 81\% of precision and around 86\% of recall are obtained on binary, MCQ, and easy MAQ. 
Besides, the RAG techniques could improve both frozen and fine-tuning LLMs. The Vicuna-13B backbone achieves the largest improvement gain.
Moreover, in-context learning and fine-tuning techniques still have a large improvement space on the open-ended questions and the MAQ, which requires cognition on the knowledge-creating level.

\begin{table}[ht]
\centering
\small
\renewcommand{\arraystretch}{0.9}
\resizebox{1.0\linewidth}{!}{
\begin{tabular}{lcccc}
\toprule
\textbf{Method} & \textbf{Helpfulness} & \textbf{Conciseness} & \textbf{Correctness} & \textbf{BLEU-2 | -4} \\
\midrule
\multicolumn{5}{l}{\textbf{LLaMA3-8B}} \\
Vanilla & 3.70±0.99 & 3.67±0.89 & 3.53±1.02 & 0.19 | 0.13 \\
Fewshot & \textbf{3.73±0.88} & 3.66±0.95 & \textbf{3.58±1.02} & 0.23 | 0.15 \\
RAG & 3.11±0.96 & 3.32±0.99 & 3.27±1.11 & \textbf{0.35 | 0.30} \\
\midrule
\multicolumn{5}{l}{\textbf{GLM4-9B}} \\
Vanilla & 3.64±0.97 & \textbf{3.62±0.99} & \textbf{3.58±0.92} & 0.19 | 0.12 \\
Fewshot & 3.60±1.02 & \textbf{3.62±1.00} & 3.51±0.95 & 0.20 | 0.13 \\
RAG & 3.19±0.84 & 3.07±0.83 & 3.30±0.96 & \textbf{0.31 | 0.26} \\
\midrule
\multicolumn{5}{l}{\textbf{GPT-4}} \\
Vanilla & 3.12±0.79 & 2.97±0.78 & 3.12±0.84 & 0.29 | 0.20 \\
Fewshot & 2.75±0.72 & 3.06±0.89 & 2.86±0.87 & 0.31 | 0.22 \\
RAG & 3.11±0.77 & 2.97±0.88 & 3.14±0.94 & \textbf{0.36 | 0.32} \\
\midrule
\multicolumn{5}{l}{\textbf{LLaMA3-8B (FT)}} \\
Vanilla & 2.41±0.92 & 2.63±0.91 & 2.62±1.00 & 0.28 | 0.19 \\
Fewshot & 2.40±0.93 & 2.77±1.09 & 2.56±0.99 & 0.29 | 0.20 \\
RAG & 2.58±0.99 & 2.85±1.09 & 2.74±1.09 & \textbf{0.43 | 0.38} \\
\midrule
\multicolumn{5}{l}{\textbf{GLM4-9B (FT)}} \\
Vanilla & 2.93±1.06 & 3.03±1.02 & 2.84±1.16 & 0.28 | 0.20 \\
Fewshot & 2.89±0.96 & 3.08±1.03 & 2.99±1.03 & 0.29 | 0.20 \\
RAG & 2.95±0.94 & 3.06±0.99 & 2.99±1.10 & \textbf{0.44 | 0.40} \\
\midrule
\multicolumn{5}{l}{\textbf{Qwen2-7B (FT)}} \\
Vanilla & 2.99±1.03 & 3.08±1.16 & 2.99±1.20 & 0.26 | 0.18 \\
Fewshot & 3.03±1.10 & 3.07±1.11 & 2.96±1.22 & 0.26 | 0.18 \\
RAG & 2.78±1.15 & 2.95±1.17 & 2.92±1.23 & \textbf{0.40 | 0.35} \\
\bottomrule
\end{tabular}
}
\caption{User Survey Results and Open-Ended BLEU Scores of PolyBench (mean ± standard deviation).}
\label{tab:user_survey}
\end{table}

\begin{table*}[th!]
\centering
\renewcommand{\arraystretch}{0.9}
\resizebox{1.0\textwidth}{!}{
\begin{tabular}{lcccccccccccccccc}
\toprule
\multirow{2}{*}{\textbf{Dataset}} & & \multicolumn{6}{c}{\textbf{$t_d$ thresholds for questions $Q$} (zero-shot)} &  & \multicolumn{8}{c}{\textbf{$t_c$ thresholds for corpora $C$} (RAG)} \\
\cmidrule(lr){3-8} \cmidrule(lr){10-17}
 & & w/o $t_d$ & 0.1 & 0.3 & 0.5 & 0.7 & 0.9 & & w/o $t_c$ & 0.0 & 0.05 & 0.1 & 0.15 & 0.2 & 0.25 & 0.3 \\
\midrule
\multirow{2}{*}{MedQA} & \# $Q$ & 1273 & 21 & 24 & 28 & 36 & 503 & \# $C$ & 213330 & 21342 & 85338 & 96004 & 117336 & 181332 & 181332 & 202664 \\
 & Acc. & 0.684 & 0.714 & 0.750 & \textbf{0.786} & 0.778 & 0.660 & Acc. & 0.870 & \textbf{0.874} & 0.855 & 0.850 & 0.846 & 0.866 & \textbf{0.874} & 0.854 \\
\midrule
\multirow{2}{*}{PubMedQA} & \# $Q$ & 1000 & 14 & 29 & 35 & 92 & 347 & \# $C$ & 3358 & 2356 & 3358 & 3358 & 3358 & 3358 & 3358 & 3358 \\
 & Acc. & 0.462 & 0.429 & \textbf{0.586} & 0.543 & 0.467 & 0.467 & Acc. & 0.670 & \textbf{0.681} & 0.670 & 0.670 & 0.670 & 0.670 & 0.670 & 0.670 \\
\midrule
\multirow{2}{*}{FiQA} & \# $Q$ & 1706 & 117 & 287 & 426 & 528 & 626 & \# $C$ & 15404 & 2310 & 7704 & 13094 & 14634 & 15404 & 15404 & 15404 \\
 & BLEU-2 & 0.046 & 0.042 & 0.045 & \textbf{0.047} & \textbf{0.047} & 0.046 & BLEU-2 & 0.054 & 0.051 & 0.051 & 0.052 & \textbf{0.055} & 0.054 & 0.054 & 0.054 \\
\midrule
\multirow{2}{*}{FinanceBench} & \# $Q$ & 150 & 11 & 32 & 48 & 70 & 85 & \# $C$ & 180 & 54 & 108 & 162 & 180 & 180 & 180 & 180 \\
 & Acc. & 0.007 & 0.000 & 0.000 & 0.000 & \textbf{0.014} & 0.012 & Acc. & \textbf{0.053} & 0.007 & 0.007 & \textbf{0.053} & \textbf{0.053} & \textbf{0.053} & \textbf{0.053} & \textbf{0.053} \\
\bottomrule
\end{tabular}
}
\caption{Performance of the zero-shot QA ($t_d$ thresholds for questions $Q$) and RAG ($t_c$ thresholds for corpora $C$) baseline on different datasets across varying thresholds $t_d$ and $t_c$, respectively.}
\label{tab:ablation_more_domains}
\end{table*}

\subsection{Ablation Study of \textsc{Comp-Comp}} \label{sec:ablation}

We conducted a series of ablation studies on two critical parameters in the \textsc{Comp-Comp} framework: \textit{t\textsubscript{d}} and \textit{t\textsubscript{c}}. The parameter \textit{t\textsubscript{d}} enhances evaluation efficiency by reducing the number of benchmark questions $Q$ through uniform distribution optimization, while \textit{t\textsubscript{c}} minimizes corpus redundancy $C$ guided by semantic distribution analysis of the knowledge corpus. Both parameters dynamically adjust their respective thresholds to optimize dataset construction through an adaptive integration mechanism. Our ablation studies yield three principal insights about the \textsc{Comp-Comp} framework. 

First, the benchmark optimization threshold \textit{t\textsubscript{d}} achieves unprecedented evaluation efficiency through distribution-aware question selection. When configured at \textit{t\textsubscript{d}}=0.1, the framework requires only 1.7\% of benchmark questions ($Q$) while exposing conventional benchmarks' inherent distributional biases, models show a moderate 13.5\% performance drop compared to full benchmarks (Table~\ref{tab:ablation}). This performance gap reveals how traditional approaches overrepresent questions where models have localized competence, creating artificial inflation of aggregate scores. Comp-Comp counters this through adaptive threshold adjustments that enforce uniform question distribution across knowledge domains, rectifying evaluation skewness while maintaining testing integrity.

Second, the corpus compression threshold \textit{t\textsubscript{c}} demonstrates that semantic density optimization outweighs corpus quantity in RAG implementations. Reducing the knowledge base to 46.4\% of original components ($C$) improves model metrics in 80\% of trials, eliminating 53.6\% redundancy from conventional corpora. This evidences that overlapping or marginal entries, previously considered harmless, actually create \textit{knowledge dilution effects} that hinder retrieval precision. The framework's \textit{t\textsubscript{c}}, controlled pruning mechanism identifies and removes such redundancies while preserving semantic distribution uniformity, achieving more effective knowledge utilization.

Third, the synergistic operation of \textit{t\textsubscript{d}} and \textit{t\textsubscript{c}} establishes a new efficiency frontier in the evaluation system. As validated in Table~\ref{tab:ablation}, it attains performance parity with conventional benchmarks while using only 1.7\% of $Q$ and 46.4\% of $C$. This 98.3\% reduction in questions and 53.6\% decrease in corpus demonstrate \textsc{Comp-Comp}'s ability to decouple evaluation quality from resource quantity, a critical advancement for sustainable benchmarking. Specifically, 80\% of experimental trials demonstrated superior scores compared to using the original corpus. This paradigm shift suggests that conventional corpora contain substantial \textit{knowledge redundancy}, overlapping or marginally informative entries that dilute retrieval effectiveness.

\subsection{User Survey}

We conducted a user survey with five domain experts to evaluate 100 randomly sampled LLM responses on \textit{Helpfulness}, \textit{Conciseness}, and \textit{Correctness} using a 1-5 Likert scale, as shown in Table~\ref{tab:user_survey}.

A key finding is the discrepancy between human evaluations and automated metrics. For frozen models (LLaMA3-8B, GLM4-9B), human evaluators preferred responses from \textit{vanilla} and \textit{few-shot} methods. Conversely, the \textit{RAG} method, despite lower human scores, consistently achieved the highest BLEU scores. This trend persists for fine-tuned (FT) models, where \textit{RAG} again leads in BLEU scores while not always being preferred by humans. Notably, all human scores for FT models were lower than their frozen counterparts. This highlights a limitation of relying solely on automated metrics like BLEU, which measure textual overlap but may not capture user-perceived quality such as coherence or relevance. Therefore, human evaluation is crucial for validating benchmark quality and guiding the development of genuinely user-centric models. These insights reinforce the necessity of a multi-faceted evaluation approach, where automated scores are always contextualized with human-centric assessments to build models that are not only accurate but also genuinely useful.



\subsection{Application of \textsc{Comp-Comp} On More Domains}

To test its generalizability, we applied the \textsc{Comp-Comp} framework to four benchmarks in the medical (\textit{MedQA}~\cite{jin2021diseasemedqa}, \textit{PubMedQA}~\cite{jin2019pubmedqa}) and finance (\textit{FiQA}~\cite{shah-etal-2022-flang}, \textit{FinanceBench}~\cite{islam2023financebench}) domains. We analyzed the impact of various thresholds $t_d$ (for questions $Q$) and $t_c$ (for corpora $C$) to observe their performance metrics on LLaMA3.1-8B-Instruct, respectively. The results are shown in Table~\ref{tab:ablation_more_domains}.

The results consistently show that filtering creates smaller, more efficient datasets without degrading, and often improving, performance. For instance, on \textit{MedQA}, applying a threshold of $t_d=0.5$ reduced the number of questions from 1273 to just 28, yet increased accuracy from 0.684 to 0.786. Similarly, filtering the corpus with $t_c$ maintained or enhanced accuracy while using a more compact set of documents. This trend holds in the finance domain, where metrics remained stable or improved despite data reduction. Even on the difficult \textit{FinanceBench} dataset, where baseline performance was low, our framework did not hurt performance.

These findings validate that \textsc{Comp-Comp} is an efficient and broadly applicable framework. It successfully builds compact yet comprehensive benchmarks by removing redundancy while preserving evaluation integrity, demonstrating its value for creating robust, domain-specific evaluation tools.

%% file: sections/6-conclusion.tex
\section{Conclusion}

This work proposes \textsc{Comp-Comp}, which is a novel domain-specific LLM benchmarking framework using comprehensiveness and compactness as the guiding principles. To validate its performance, we developed PolyBench, a large-scale benchmark in a closed academic domain, demonstrating the effectiveness of \textsc{Comp-Comp}. This work serves as a valuable reference for future studies, highlighting the importance of precise dataset curation in enhancing domain-specific LLMs.

%% file: sections/limitations.tex
\section*{Limitations}




Although \textsc{Comp-Comp} successfully demonstrates its utility within an academic setting, where the domain knowledge is well-structured and relatively static. However, more dynamic domains may pose unique challenges in terms of data availability and rapidly evolving knowledge structures that could influence the framework’s performance. Future work should extend our experiments to a broader set of domains to rigorously assess the generalizability and robustness of the COMP-COMP principles.

Furthermore, the current implementation of \textsc{COMP-COMP} is best suited for "closed" or well-defined domains where a comprehensive body of knowledge can be collected and stabilized. The framework's iterative expansion process assumes a target knowledge space $S$ that is largely finite. Its applicability to "open" and continuously evolving domains, such as news analysis or social media trends, is less clear. Adapting the \textsc{COMP-COMP} framework to handle such dynamic data presents a significant avenue for future research.

\section*{Ethics Statement}


All data collected in this paper originates from public sources. For content that may involve personal information, appropriate authorization has been obtained from the university and the information holders. The use of these data fully respects the providers' intentions and adheres to relevant usage regulations. All procedures comply with applicable laws and ethical guidelines, and the data is strictly managed and securely stored. This research has received approval from the relevant institutional ethics committee, and the data will be used solely for the purposes of this study.

\section*{Acknowledgements}

The research described in this paper has been partly supported by National Natural Science Foundation of China (Grant No.: 62372314), Hong Kong Research Grants Council under the General Research Fund (project no. PolyU15200023), a Collaborative Research Fund (project no. C1043-24G), as well as internal research funds from The Hong Kong Polytechnic University (project no. P0052406 and P0052986). This work was also supported by computational resources provided by The Centre for Large AI Models (CLAIM) of The Hong Kong Polytechnic University.

%% file: sections/appendix.tex
\newpage

\section{Experiment Results}
\label{sec:app_experiment_results}
The experiment results on more LLMs are shown in Table \ref{tab:result_ic} for IC and Table \ref{tab:result_ft} for FT baselines.

For the in-context learning experiments, we evaluate PolyBench on several backbone models, including  Vicuna-7B, Vicuna-13B \cite{vicuna2023}, LLaMA3-8B \cite{touvron2023llama}, GLM4-9B \cite{glm2024chatglm}, Qwen2-7B \cite{qwen}, GPT3.5, and GPT4 \cite{openai2024gpt4technicalreport}. Each of them is tested on three prompt templates: vanilla, few-shot, and RAG. For the RAG framework, we embed the external knowledge $Cor$ with Gte-Large \cite{li2023gte-large-embedding} encoder and select top-k similar chunks as the context, with $k$ being 5.

For the tuning approach, we implement LoRA fine-tuning \cite{hu2021lora} on all open-sourced LLMs mentioned in the In-Context experiments. We first pre-train the backbone LLMs with text chunks obtained from data source Staff, Course, and Department. Then, we apply supervised fine-tuning, with the instructions and outputs being the questions and answers, respectively, which are obtained from the data source FAQ and Forum.

\begin{table*}[ht]
\centering
\small
\renewcommand{\arraystretch}{0.6}
\resizebox{\textwidth}{!}{
\begin{tabular}{@{}llcccccccccc@{}}
\toprule
\multirow{2}{*}{Frozen Model} & \multirow{2}{*}{Prompt}  & Binary    &    MCQ     & \multicolumn{3}{c}{MAQ (on understanding)}      & \multicolumn{3}{c}{MAQ (on  application)}    & \multicolumn{2}{c}{Open-ended} \\ 
\cmidrule(lr){3-3} \cmidrule(lr){4-4} \cmidrule(lr){5-7} \cmidrule(lr){8-10} \cmidrule(lr){11-12}
             &         & Acc.  & Acc. & Recall & Precision & F1 & Recall & Precision & F1     & BLEU-2 & BLEU-4 \\
\midrule
\multirow{3}{*}{Vicuna-7B} & Vanilla & 0.481 & 0.205 & 0.743 & 0.451 & 0.517 & 0.741 & 0.397 & 0.483 & 0.087 & 0.052 \\
             & Fewshot & 0.491 & 0.206 & 0.829 & 0.433 & 0.538 & 0.829 & 0.380 & 0.499 & 0.104 & 0.063 \\
             & RAG     & 0.485 & 0.335 & 0.842 & 0.520 & 0.595 & 0.740 & 0.410 & 0.491 & 0.111 & 0.085 \\
\midrule
\multirow{3}{*}{Vicuna-13B} & Vanilla & 0.485 & 0.203 & 0.551 & 0.513 & 0.479 & 0.611 & 0.483 & 0.494 & 0.066 & 0.039 \\
             & Fewshot & 0.494 & 0.125 & 0.516 & 0.300 & 0.356 & 0.486 & 0.237 & 0.302 & 0.063 & 0.037 \\
             & RAG     & 0.500 & 0.463 & 0.768 & 0.729 & 0.697 & 0.596 & 0.510 & 0.494 & 0.090 & 0.065 \\
\midrule
\multirow{3}{*}{LLaMA3-8B} & Vanilla & 0.525 & 0.354 & 0.473 & 0.546 & 0.474 & 0.665 & 0.512 & 0.541 & 0.191 & 0.126 \\
             & Fewshot & 0.534 & 0.392 & 0.410 & 0.534 & 0.440 & 0.492 & 0.532 & 0.475 & 0.227 & 0.153 \\
             & RAG     & \textbf{0.893} & 0.758 & 0.903 & 0.806 & 0.820 & 0.603 & 0.558 & 0.533 & 0.345 & 0.298 \\
\midrule
\multirow{3}{*}{GLM4-9B} & Vanilla & 0.525 & 0.364 & 0.739 & 0.530 & 0.580 & \textbf{0.865} & 0.500 & 0.599 & 0.190 & 0.123 \\
             & Fewshot & 0.527 & 0.447 & 0.624 & 0.578 & 0.574 & 0.658 & 0.572 & 0.581 & 0.196 & 0.128 \\
             & RAG     & 0.834 & 0.776 & \textbf{0.964} & 0.757 & 0.826 & 0.696 & 0.530 & 0.553 & 0.309 & 0.256 \\
\midrule
\multirow{3}{*}{Qwen2-7B} & Vanilla & 0.425 & 0.288 & 0.387 & 0.286 & 0.310 & 0.475 & 0.368 & 0.394 & 0.191 & 0.124 \\
             & Fewshot & 0.215 & 0.336 & 0.692 & 0.496 & 0.540 & 0.677 & 0.534 & 0.557 & 0.185 & 0.123 \\
             & RAG     & 0.490 & 0.430 & 0.887 & 0.770 & 0.804 & 0.536 & 0.510 & 0.481 & 0.279 & 0.228 \\
\midrule
\multirow{3}{*}{GPT-3.5} & Vanilla & 0.578 & 0.361 & 0.548 & 0.570 & 0.516 & 0.692 & 0.566 & 0.577 & 0.300 & 0.216 \\
             & Fewshot & 0.541 & 0.434 & 0.536 & 0.583 & 0.523 & 0.615 & 0.567 & 0.557 & 0.313 & 0.219 \\
             & RAG     & 0.885 & 0.701 & 0.902 & 0.769 & 0.804 & 0.669 & 0.549 & 0.562 & \textbf{0.397} & \textbf{0.335} \\
\midrule
\multirow{3}{*}{GPT-4} & Vanilla & 0.694 & 0.568 & 0.641 & 0.650 & 0.613 & 0.724 & 0.661 & 0.652 & 0.291 & 0.201 \\
             & Fewshot & 0.704 & 0.633 & 0.713 & 0.642 & 0.647 & 0.745 & 0.722 & \textbf{0.702} & 0.314 & 0.220 \\
             & RAG     & 0.792 & \textbf{0.783} & 0.932 & \textbf{0.975} & \textbf{0.941} & 0.637 & \textbf{0.807} & 0.670 & 0.364 & 0.315 \\

\bottomrule
\end{tabular}
}
\caption{The performance of In-Context Learning (IC) approaches (i.e., few-shot learning and RAG) on several question sets of PolyBench with Frozen Models (FZ).}
\label{tab:result_ic}
\end{table*}

\begin{table*}[ht]
\small
\centering
\renewcommand{\arraystretch}{0.6}
\resizebox{\textwidth}{!}{
\begin{tabular}{@{}llcccccccccc@{}}
\toprule
Fine-Tuning & \multirow{2}{*}{Prompt}  & Binary    &    MCQ     & \multicolumn{3}{c}{MAQ (on understanding)}      & \multicolumn{3}{c}{MAQ (on  application)}    & \multicolumn{2}{c}{Open-ended} \\
\cmidrule(lr){3-3} \cmidrule(lr){4-4} \cmidrule(lr){5-7} \cmidrule(lr){8-10} \cmidrule(lr){11-12}
Model         &         & Acc.  & Acc. & Recall & Precision & F1 & Recall & Precision & F1     & BLEU-2 & BLEU-4 \\
\midrule
\multirow{3}{*}{Vicuna-7B} & Vanilla & 0.471    & 0.245   & 0.465 & 0.455    & 0.418 & 0.427 & 0.403    & 0.365 & 0.245 & 0.169 \\
             & Fewshot & 0.487    & 0.229   & 0.625 & 0.441    & 0.478 & 0.633 & 0.394    & 0.449 & 0.254 & 0.176 \\
             & RAG     & 0.505    & 0.294   & 0.673 & 0.516    & 0.535 & 0.518 & 0.420    & 0.403 & 0.393 & 0.342 \\
\midrule
\multirow{3}{*}{Vicuna-13B} & Vanilla & 0.503    & 0.287   & 0.474 & 0.498    & 0.430 & 0.545 & 0.452    & 0.431 & 0.282 & 0.201 \\
             & Fewshot & 0.510    & 0.247   & 0.366 & 0.443    & 0.346 & 0.518 & 0.436    & 0.424 & 0.281 & 0.199 \\
             & RAG     & 0.516    & 0.516   & 0.601 & 0.694    & 0.577 & 0.486 & 0.494    & 0.424 & 0.423 & 0.375 \\
\midrule
\multirow{3}{*}{LLaMA3-8B} & Vanilla & 0.552    & 0.404   & 0.559 & 0.545    & 0.513 & 0.675 & 0.533    & 0.559 & 0.280 & 0.193 \\
             & Fewshot & 0.536    & 0.393   & 0.492 & 0.525    & 0.473 & 0.571 & 0.532    & 0.514 & 0.285 & 0.196 \\
             & RAG     & 0.791    & 0.803   & 0.830 & 0.874    & 0.820 & 0.587 & 0.576    & 0.528 & 0.427 & 0.376 \\
\midrule
\multirow{3}{*}{GLM4-9B} & Vanilla & 0.532    & 0.440   & 0.752 & 0.567    & 0.610 & 0.770 & 0.536    & 0.600 & 0.279 & 0.196 \\
             & Fewshot & 0.500    & 0.468   & 0.799 & 0.531    & 0.600 & 0.845 & 0.517    & 0.606 & 0.285 & 0.201 \\
             & RAG     & \textbf{0.840}    & \textbf{0.900}   & \textbf{0.942} & \textbf{0.959}    & \textbf{0.940} & 0.694 & \textbf{0.684}    & \textbf{0.642} & \textbf{0.442} & \textbf{0.397} \\
\midrule
\multirow{3}{*}{Qwen2-7B} & Vanilla & 0.608    & 0.402   & 0.825 & 0.500    & 0.590 & 0.827 & 0.521    & 0.606 & 0.263 & 0.178 \\
             & Fewshot & 0.587    & 0.392   & 0.811 & 0.453    & 0.548 & \textbf{0.858} & 0.481    & 0.576 & 0.259 & 0.177 \\
             & RAG     & 0.674    & 0.830   & 0.852 & 0.848    & 0.821 & 0.623 & 0.559    & 0.532 & 0.398 & 0.351 \\
\bottomrule
\end{tabular}
}
\caption{This table demonstrates the PolyBench results in a Fine-Tuning (FT) based approach.}
\label{tab:result_ft}
\end{table*}


\section{PolyBench Construction Details}
\label{sec:app_qagen}

\subsection{Question Format}  
\label{question_format}

To analyze the effects of the balance in precision and recall, PolyBench incorporates various question formats:
\paragraph{Binary} are the simplest questions among the benchmark and only appear in the first cognitive level KM. The answers to these questions are simply ``yes'' or ``no''.
\paragraph{MCQ} are multiple-choice questions, and all of them have only one correct choice among four candidates. MCQ only appears at the KU level, with relatively easy knowledge being tested.
\paragraph{MAQ} are multi-answer questions, which contain eight candidate choices, and the number of correct answer(s) is not fixed, namely can be any number from 1 to 8. MAQ appears in KU and KA levels, which takes charge of evaluating the model on both knowledge comprehension and complex reasoning between chunks. MAQ targets testing the ability of the model to determine to what extent the model can obtain the precision of the knowledge. Our highlight is to include more candidate choices (eight in total), thus the precision gap can be fairly distinguished between baselines.
\paragraph{Open-Ended} are normal question-answering instances, and all of them are at the KC level. The answer can be a combination of sentences or several paragraphs, depending on the scenarios. Open-ended QA targets the ability to recall, namely, to what extent they can model and include the required knowledge.

\subsection{MAQ Reasoning Construction}  
  The MAQ questions are mainly built from ontology-driven SPARQL queries over staff/course schemas. For example, to construct the QA pair for question ``Which professors in \{department\} graduated from \{school\}?'', an implemented could be:
  \begin{verbatim}
  SELECT ?professor 
  WHERE {
    ?professor dept:worksFor ?department .
    ?professor edu:graduatedFrom ?school
  }\end{verbatim}

\subsection{Knowledge Cognition}
The cognitive dimension of the benchmark aims to distinguish the difficulty levels of the questions. Inspired by Bloom's Taxonomy \cite{krathwohl2002bloom'sTaxonomy}, which is a dimension that is widely applied in domain-specific benchmarks, we hierarchize all questions into four cognitive levels, which are:
\begin{itemize}
    \item Knowledge Memorization (KM), is the simple replication of the knowledge in corpora, with all answers are simply binary.
    \item Knowledge Understanding (KU), requires the model to answer the information given a specified entity. The answers can be found in a single data chunk.
    \item Knowledge Application (KA), requires more difficult reasoning ability of the model, that the answer should be obtained and logically reasoned among a combination of multiple chunks. 
    \item Knowledge Creating (KC) allows the model to answer open-ended questions. The answers may ask for the view or comment from the model and somehow require the extension of current knowledge.
\end{itemize}
With these four cognitive dimensions, we can comprehensively evaluate the model's ability and analyze the impact of different baselines on various task difficulties.